# D3FNet: A Differential Attention Fusion Network for Fine-Grained Road Structure Extraction in Remote Perception Systems

Chang Liu[1,2]*   Yang Xu[1]   Tamas Sziranyi[1,2]

[1]Department of Networked Systems and Services, Faculty of Electrical Engineering and Informatics,
Budapest University of Technology and Economics, Budapest, Hungary

[2]Machine Perception Research Laboratory,
HUN-REN Institute for Computer Science and Control (HUN-REN SZTAKI), Budapest, Hungary

`liu.chang@sztaki.hun-ren.hu, xuy@edu.bme.hu, sziranyi.tamas@sztaki.hun-ren.hu`

## Abstract

*Extracting narrow roads from high-resolution remote sensing imagery remains a significant challenge due to their limited width, fragmented topology, and frequent occlusions. To address these issues, we propose **D3FNet**, a Dilated Dual-Stream Differential Attention Fusion Network designed for fine-grained road structure segmentation in remote perception systems. Built upon the encoder–decoder backbone of D-LinkNet, D3FNet introduces three key innovations: (1) a Differential Attention Dilation Extraction (DADE) module that enhances subtle road features while suppressing background noise at the bottleneck; (2) a Dual-stream Decoding Fusion Mechanism (DDFM) that integrates original and attention-modulated features to balance spatial precision with semantic context; and (3) a multi-scale dilation strategy (rates 1, 3, 5, 9) that mitigates gridding artifacts and improves continuity in narrow road prediction. Unlike conventional models that overfit to generic road widths, D3FNet specifically targets fine-grained, occluded, and low-contrast road segments. Extensive experiments on the DeepGlobe and CHN6-CUG benchmarks show that D3FNet achieves superior IoU and recall on challenging road regions, outperforming state-of-the-art baselines. Ablation studies further verify the complementary synergy of attention-guided encoding and dual-path decoding. These results confirm D3FNet as a robust solution for fine-grained narrow road extraction in complex remote and cooperative perception scenarios.*

## 1. Introduction

In cooperative autonomous driving systems, remote perception is increasingly recognized as a critical component for global scene understanding, high-definition (HD) map generation, and robust navigation in unstructured environments. While most V2X-based systems rely on vehicle-mounted and roadside sensors, aerial and satellite imagery extends perception beyond local line-of-sight, offering global context for route-level decision-making. In scenarios where ground sensors are limited—such as rural areas, disaster zones, or GPS-denied environments—remote sensing imagery provides vital input for collaborative planning and hazard-aware routing [12, 16].

Among remote perception tasks, fine-grained road structure extraction from high-resolution satellite imagery has become particularly important. Accurate narrow road segmentation supports topological map construction, enables efficient routing, and ensures safety-critical decision-making in autonomous systems [13]. However, this task remains highly challenging due to multiple factors: (1) the inherently narrow width of roads relative to background objects, (2) frequent topological disconnections caused by occlusions (e.g., trees, buildings, shadows), and (3) low contrast and high intra-class variability across regions, seasons, and sensors. These challenges significantly degrade the performance of conventional segmentation models.

Existing methods, such as ResUNet [28] and D-LinkNet [30], have made progress by leveraging encoder–decoder architectures and skip connections. However, they often exhibit weak localization for thin structures and are prone to losing continuity under occlusion or visual ambiguity. Moreover, many of these models are optimized for wide, prominent roads and fail to generalize to finer structures essential for real-time planning and safe navigation in cooperative driving systems.

To address these limitations, we propose **D3FNet**—a *Dilated Dual-Stream Differential Attention Fusion Network* designed specifically for fine-grained road extraction from remote sensing imagery. Built upon the D-LinkNet [30]

---

*Corresponding author: changliu@hit.bme.hu, changliu@sztaki.hu

backbone, D3FNet introduces three key innovations:

- A **Differential Attention Dilation Extraction (DADE)** module, which enhances subtle road features while effectively suppressing cluttered backgrounds;
- A **Dual-stream Decoding Fusion Mechanism (DDFM)** that fuses original and attention-refined features to jointly capture semantic consistency and spatial localization;
- An **optimized multi-scale dilation strategy** to improve contextual representation while avoiding gridding artifacts.

We evaluate D3FNet on two challenging datasets—**DeepGlobe** [7] and **CHN6-CUG** [32]—covering diverse geographies and road morphologies. Our model consistently outperforms state-of-the-art baselines in terms of IoU, Recall, and F1-score, especially on narrow and fragmented road segments. These results highlight the effectiveness of D3FNet for remote narrow road perception tasks that directly support cooperative autonomous navigation.

**Our main contributions are summarized as follows:**
- We propose D3FNet, a novel architecture for narrow road extraction, combining differential attention [26] and dual-stream decoding into the D-LinkNet [30] framework.
- We design a DADE module and DDFM structure that improve feature discrimination and preserve spatial continuity for narrow, occluded roads.
- We demonstrate state-of-the-art performance on two benchmark road extraction datasets, validating the effectiveness and generalization ability of our method for cooperative driving scenarios.

The remainder of this paper is organized as follows. Section 2 reviews related work on road extraction from remote sensing imagery. Section 3 details the proposed D3FNet architecture, including the DADE module and dual-stream decoding fusion mechanism. Section 4 presents the experimental setup and evaluates performance on benchmark datasets. Finally, Section 5 concludes the paper and discusses potential future research directions.

## 2. Related Work

### 2.1. Road Extraction from Remote Sensing Imagery

Road extraction is a fundamental task in remote sensing image analysis with wide applications in intelligent transportation, urban planning, and disaster response [13, 19]. High-precision road networks support the construction of detailed maps and provide essential information for autonomous driving and emergency management systems [15]. However, extracting roads from remote sensing imagery faces significant challenges, including the narrow width of roads compared to background objects, frequent topological discontinuities caused by occlusions such as trees, buildings, and shadows, and low contrast with high intra-class variability across different regions, seasons, and sensors [17].

Classic approaches for road extraction often rely on deep learning-based semantic segmentation networks, such as U-Net [20] and its improved variant ResUNet [28], which utilize encoder-decoder architectures with skip connections to fuse multi-scale features and enhance boundary localization. D-LinkNet [30] builds upon this by incorporating dilated convolutions to expand the receptive field, effectively capturing long-range dependencies. DeepLabV3+ [5] further improves semantic feature extraction through Atrous Spatial Pyramid Pooling (ASPP), becoming a strong baseline for road extraction tasks in remote sensing.

More recently, advanced models have emerged to address the complex topological and occlusion challenges in road extraction. For instance, using a Swin Transformer-based UNet architecture [9], the model effectively captures both global and local features of road images through self-attention mechanisms, thereby improving the accuracy and structural connectivity of road extraction. Graph neural network (GNN)-based methods [3] explicitly model road nodes and edges to optimize the completeness and structural integrity of road networks. Additionally, multi-scale and multi-task learning strategies [14] have been widely adopted to simultaneously improve fine boundary extraction and semantic classification of roads. In particular, several recent works have focused specifically on the challenges of narrow road extraction. SWGE-Net and MSIF-Net [29] introduce edge-guided and multi-scale integration modules, respectively, to better capture subtle features of narrow road structures in high-resolution imagery. These methods demonstrate improved performance on narrow roads but still suffer from limitations in maintaining structural continuity under severe occlusion or ambiguous visual contexts.

Despite these advances, most existing methods focus primarily on wide and continuous main roads, and their performance on narrow, fragmented, and heavily occluded road segments remains limited [1, 13]. In practical scenarios where image resolution is limited and environments are highly variable, these models often fail to preserve road continuity and integrity, which restricts their applicability in high-precision tasks such as autonomous navigation and disaster rescue. Therefore, there is an urgent need for innovative approaches tailored to narrow road extraction under challenging remote sensing conditions.

### 2.2. Attention Mechanisms and Feature Fusion for Narrow Road Semantic Segmentation

Narrow road structures in remote sensing images typically appear as thin and elongated objects with low contrast, making them highly susceptible to being overwhelmed by complex background information. Furthermore, occlusions caused by buildings, vegetation, and shadows often lead to topological discontinuities in road extraction [25]. These challenges render the effective fusion of spatial and semantic information critical for precise segmentation.

Attention mechanisms have been shown to significantly enhance the continuity and discriminability of narrow road extraction by guiding the network to focus on key regions and suppress background noise. Channel-wise attention mechanisms, such as SE-Net [11] and CBAM [23], improve the semantic representation by modeling the importance of feature channels but are limited in preserving spatial details. Spatial attention and dual attention mechanisms, exemplified by DANet [8], simultaneously leverage spatial and channel information to substantially improve structural continuity, making them well-suited for fine-grained and occluded road detection tasks. Notably, differential attention mechanisms [26] compare features between adjacent encoding layers to effectively capture subtle and fragmented road segments, thereby enhancing sensitivity to low-contrast and occluded regions.

Regarding feature fusion between high-resolution shallow spatial features and deep semantic features, we design a Dual-stream Decoder with Fusion Mechanism (DDFM). This module performs parallel processing and fusion of shallow and deep features, using attention-guided modulation of deep semantic features by shallow spatial features. Such design enhances boundary localization accuracy while maintaining semantic consistency and topological integrity [21, 24]. To address the gridding artifacts common in dilated convolution-based multi-scale modeling, we propose the Differential Attention Dilation Extraction (DADE) module. DADE employs a carefully selected set of dilation rates $\{1, 3, 5, 9\}$ to effectively cover diverse receptive field scales and mitigate uneven sampling effects [5, 27]. ombined with differential attention, the DADE module dynamically emphasizes local structural variations, boosting recognition of fragmented and low-contrast roads and significantly improving the continuity and accuracy of narrow road extraction.

In summary, current research incorporating differential attention mechanisms and dual-stream fusion strategies, together with a well-designed multi-scale dilation strategy, effectively overcomes limitations in detail preservation, occlusion recovery, and scale variation in complex remote sensing imagery, thus providing robust support for accurate narrow road semantic segmentation.

## 3. Methodology

### 3.1. Overview of D3FNet Architecture

D3FNet is a novel architecture designed for narrow road extraction, introducing a differential attention mechanism [26] to enhance feature discrimination. As shown in Figure 1, D3FNet builds upon the encoder-decoder structure of D-LinkNet [30]. The input satellite image is first encoded by a pre-trained ResNet34 backbone [10], which reduces spatial resolution while extracting multi-level semantic features. These features are then processed by the Differential Attention Dilation Extraction (DADE) module in the network center, which integrates four dilated convolutions with dilation rates of 1, 3, 5, and 9 to capture multi-scale structures. Each dilated convolutional layer is followed by a multi-head differential attention module that suppresses background noise through the subtraction of two independently computed attention maps, thereby enhancing the task-relevant features. To prevent gradient interference between dilation and attention during training, a Dual-stream Decoding Fusion Mechanism (DDFM) is adopted. The DADE output is split into two streams: a structural stream enhanced by encoder skip connections and an attention stream that focuses on global semantics. Both streams are decoded independently through up-sampling blocks and then fused to generate the final high-resolution binary segmentation map indicating road areas.

### 3.2. Differential Attention Dilation Extraction (DADE) Module

Inspired by the central module of D-LinkNet, we design the Differential Attention Dilation Extraction (DADE) module (Fig. 2), which consists of four cascaded dilated convolutional layers, each followed by a parallel multi-head differential attention mechanism to enhance feature discriminability. The differential attention mechanism, derived from the Differential Transformer [26] architecture, is an improved attention method aimed at reducing the excessive focus on irrelevant context often seen in traditional transformers. Its core idea is to suppress shared noisy activations by computing the element-wise difference between two independently generated attention maps, thereby emphasizing road-relevant features. Specifically, the input is first processed by an embedding layer, and the resulting feature representations are then split into two independent sets of query (Q) and key (K) matrices, which are used to compute the corresponding attention weight matrices. These matrices are then combined through a weighted subtraction to produce the differential attention weights. The resulting weights are applied to the value (V) matrix.

Mathematically, the input features $X \in \mathbb{R}^{b \times n \times c}$ are projected into query, key, and value tensors $Q, K, V \in \mathbb{R}^{b \times n \times 2d}$ by linear projections with weight matrices $W_Q, W_K, W_V \in \mathbb{R}^{c \times 2d}$, respectively. Then, $Q$ and $K$ are split along the last dimension into two parts of size $d$ as follows:

$$\begin{aligned} Q &= XW_Q, \quad Q_1, Q_2 = \text{split}(Q), \\ K &= XW_K, \quad K_1, K_2 = \text{split}(K), \\ V &= XW_V. \end{aligned}$$

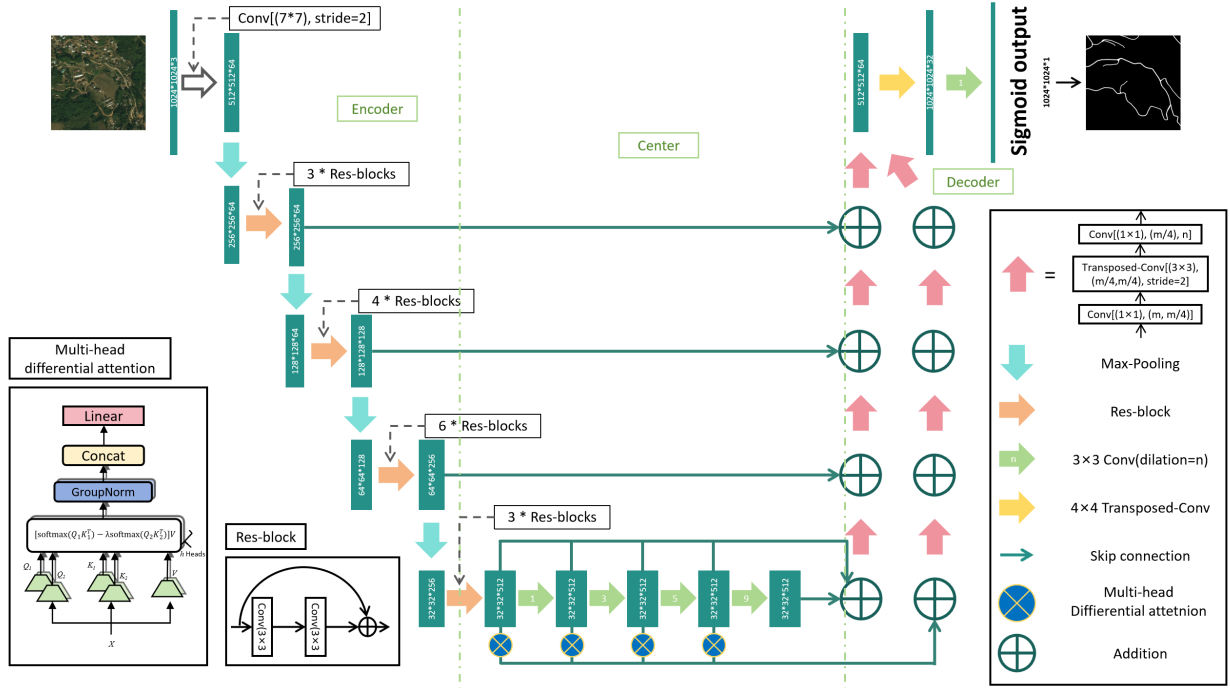

Figure 1. D3FNet architecture. Each green rectangular block represents a multi-channel feature map. In the encoder, D3FNet adopts ResNet34 to extract hierarchical spatial features. The center part includes the Differential Attention and Dilated Extraction (DADE) module, which consists of four cascaded dilated convolutional layers, each followed by a parallel multi-head differential attention mechanism. In the decoder, based on LinkNet, a Dual-stream Decoding Fusion Mechanism (DDFM) is introduced, where the structural stream from dilated convolutions receives skip connections from the encoder, while the attention stream focuses on semantic refinement without direct encoder connections.

The attention maps are:

$$A_1 = \text{softmax}\left(\frac{Q_1 K_1^T}{\sqrt{d}}\right), \quad A_2 = \text{softmax}\left(\frac{Q_2 K_2^T}{\sqrt{d}}\right).$$

The final differential attention output is computed as:

$$\text{DiffAttention}((X) = (A_1 - \lambda A_2)V$$

where $\lambda \in [0, 1]$ controls the suppression strength. This differential mechanism effectively acts as a filter to cancel redundant or background attention and highlight road-related features.

This mechanism, similar to a differential amplifier, cancels common noise and enhances sensitivity to key features. Dilated convolutions capture multi-scale spatial structures, while differential attention suppresses background noise by computing differences between two independent attention maps. By integrating dilated convolutions and differential attention in parallel, the DADE module effectively captures multi-scale spatial context while suppressing background noise, resulting in enhanced semantic discrimination and finer road detail extraction.

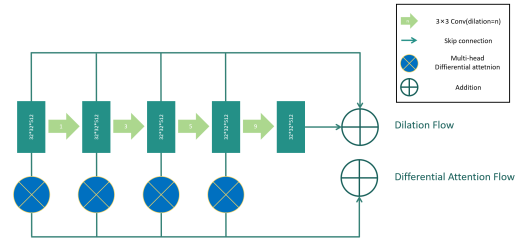

Figure 2. Differential attention dilation extraction module.

### 3.3. Dual-Stream Decoding Fusion Mechanism (DDFM)

To mitigate gradient interference between dilated convolutions and differential attention during backpropagation—which may destabilize training and reduce model performance—this study introduces a Dual-stream Decoding Fusion Mechanism (DDFM) in the decoder. This mechanism integrates structural features and attention-enhanced contextual information extracted from the central module. As shown in Fig. 3, the output is split into two parallel streams: one carries structural features from dilated convolutions, and the other aggregates contextual cues refined by multi-head differential attention.

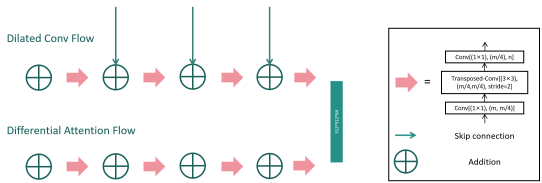

Figure 3. Dual-stream decoding fusion mechanism.

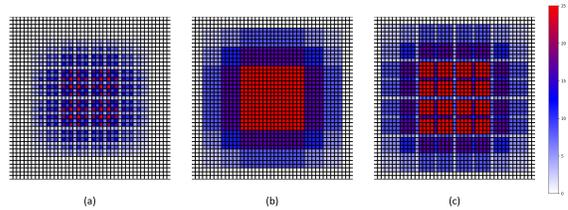

Figure 4. Comparison of receptive fields with different dilation rates: (a), (b), and (c) represent the final receptive fields with different dilation rates of (1, 2, 4, 8), (1, 3, 5, 9), and (1, 3, 5, 10).

During decoding, skip connections from the encoder are added only to the structural stream to preserve fine-grained spatial details, while the attention stream focuses on global semantics, avoiding basic-level noise. To prevent gradient entanglement between the structural and semantic paths during training, we decouple them via DDFM. In preliminary experiments, directly merging the two paths led to unstable loss oscillations and degraded validation IoU, highlighting the necessity of stream decoupling. Fig. 3 highlights this decoupling by removing skip-connections from the attention stream. Each decoding path contains four blocks that reduce channel dimensions via 1×1 convolutions, followed by batch normalization, ReLU activation, and transposed convolutions for upsampling and refinement. The outputs of both streams are then concatenated and passed through a fusion layer to produce a high-resolution road segmentation map. This dual-stream design effectively combines structural and contextual features, improving segmentation accuracy and robustness.

### 3.4. Multi-Scale Dilation Strategy

Dilated convolution enables multi-scale context aggregation by enlarging the receptive field without sacrificing spatial resolution or increasing parameter complexity. In the proposed Differential Attention Dilation Extraction (DADE) module, we employ four cascaded dilated convolutional layers. The original D-LinkNet uses dilation rates of (1, 2, 4, 8), which can cause the "gridding effect"—a sparse, checkerboard-like receptive field—due to shared factors among dilation rates. To mitigate this, the Hybrid Dilated Convolution (HDC) strategy [22] suggests selecting dilation rates without common factors. We compare three dilation rate configurations—(1, 2, 4, 8), (1, 3, 5, 9), and (1, 3, 5, 10)—as illustrated in Fig. 4. In these visualizations, deeper colors represent higher coverage frequency. Both (1, 3, 5, 9) and (1, 3, 5, 10) reduce the gridding effect compared to the original setting. However, (1, 3, 5, 9) demonstrates superior uniformity and continuity in receptive field distribution. Based on these observations, we adopt (1, 3, 5, 9) as the dilation configuration in our model. This design decision plays a pivotal role in enhancing the continuity of road predictions and complements the attention-guided encoding and dual-stream decoding processes described earlier.

## 4. Experiments

We train our model on the DeepGlobe [7] and CHN6-CUG [32] datasets and evaluate its performance using standard metrics, including Intersection over Union (IoU), Precision, Recall, and F1-score. The results are then compared against our baseline model D-LinkNet [30], as well as several recent road extraction approaches to validate the effectiveness of D3FNet.

### 4.1. Datasets

We utilize two publicly available remote sensing datasets, DeepGlobe [7] and CHN6-CUG [32], to train and evaluate our model. The DeepGlobe dataset consists of high-resolution (0.5 m/pixel) RGB satellite images from regions in Thailand, Indonesia, and India. It contains 8,570 images of size $1024 \times 1024$ pixels, among which 6,226 have pixel-level road annotations. In our experiments, 95% of these annotated images are used for training, and the remaining 5% for testing. To further evaluate the generalizability of the model, we employ the CHN6-CUG dataset, which also provides RGB satellite imagery at a resolution of 0.5 m/pixel. This dataset comprises 4,511 images of size $512 \times 512$ pixels, collected from six cities across China and covering diverse road types including urban streets, highways, rural roads, and railways. We split this dataset into 80% for training and 20% for testing.

These datasets together offer a diverse set of road environments and imaging conditions, enabling comprehensive evaluation of the proposed method's robustness and applicability.

### 4.2. Evaluation Metrics

Road extraction is treated as a pixel-wise binary classification task. Model performance is evaluated based on the confusion matrix, which categorizes each pixel prediction into true positives (TP), true negatives (TN), false positives (FP), and false negatives (FN). We employ four commonly used metrics to comprehensively assess the segmentation quality: Intersection over Union (IoU), Precision, Recall, and F1-score. Specifically,

- **IoU** measures the overlap between predicted and ground truth road regions,
- **Precision** indicates the proportion of correctly predicted road pixels among all pixels predicted as roads,
- **Recall** reflects the model's ability to identify actual road pixels, and
- **F1-score** balances Precision and Recall, making it particularly useful in cases of class imbalance.

Together, these metrics provide a thorough understanding of the model's accuracy, completeness, and robustness in road segmentation tasks.

### 4.3. Implementation Details

All experiments were conducted on a system running Windows 11, equipped with a 14th Gen Intel Core i7-14650HX processor, 16 GB of RAM, and an NVIDIA GeForce RTX 4060 Laptop GPU with 8 GB of VRAM. The software environment included Python 3.12.4, CUDA 12.5, and PyTorch 2.4. We adopted the ReLU activation function and the Adam optimizer, using a learning rate of $1 \times 10^{-4}$ and a batch size of 4 to balance training efficiency and memory consumption. The model was independently trained and evaluated on the DeepGlobe [7] and CHN6-CUG [32] datasets to comprehensively assess its performance. Specifically, training on DeepGlobe was conducted for 50 epochs, while CHN6-CUG training spanned 150 epochs to ensure convergence and robustness across different data domains.

### 4.4. Experimental Results

#### 4.4.1. Quantitative Results

Table 1 presents a quantitative comparison of D3FNet with several widely used road extraction models on the DeepGlobe dataset, including U-Net [28], U-Net++ [31], DeepLabV3+ [5], and D-LinkNet34 [30]. The results of these models on the DeepGlobe dataset are reported by Zhao et al. [29].

In terms of key evaluation metrics, including Intersection over Union (IoU), F1-score, Precision, and Recall, D3FNet demonstrates clear improvements in road detection coverage and recall capability. Specifically, D3FNet achieves an IoU of 63.18%, noticeably higher than D-LinkNet's 60.39%, and an F1-score of 75.95%, slightly surpassing D-LinkNet34's 75.30%. Notably, D3FNet attains a Recall of 82.78%, a substantial improvement over D-LinkNet's 66.76%, indicating a stronger ability to identify road regions. However, D3FNet's Precision is 73.79%, which is significantly lower than D-LinkNet's 86.36%, and also lower than DeepLabV3+'s 90.10%, the highest among all models. This drop in precision suggests a higher false positive rate, which we attribute in part to ground truth annotation errors: some narrow or degraded road segments correctly identified by the model are missing in the ground truth, and thus counted as false positives (see qualitative examples in Fig. 5).

Table 1. Quantitative comparison on DeepGlobe dataset.

| Model | Backbone | IoU (%) | F1-score (%) | Precision (%) | Recall (%) |
|---|---|---|---|---|---|
| U-net [28] | None | 53.99 | 70.12 | 85.32 | 59.72 |
| U-net++ [31] | None | 39.20 | 56.32 | 84.86 | 42.15 |
| DeepLabV3+ [5] | Xception | 44.23 | 61.34 | **90.10** | 46.49 |
| D-linkNet34 [30] | ResNet34 | 60.39 | 75.30 | 86.36 | 66.76 |
| **Ours (D3FNet)** | ResNet34 | **63.18** | **75.95** | 73.79 | **82.78** |

Table 2. Quantitative comparison on CHN6-CUG dataset

| Model | Backbone | IoU (%) | F1 (%) | Prec. (%) | Rec. (%) |
|---|---|---|---|---|---|
| U-net [28] | None | 48.57 | 63.77 | 68.42 | 59.72 |
| U-net++ [31] | None | 47.38 | 63.91 | 68.33 | 60.02 |
| DeepLabV3+ [5] | Xception | 52.04 | 65.20 | 72.24 | 59.41 |
| D-linkNet34 [30] | ResNet34 | 57.56 | 69.26 | 72.61 | 66.21 |
| TransUNet [4] | - | 31.74 | 48.18 | 69.84 | 36.78 |
| Swin Transformer [18] | Swin-T | 34.10 | 50.86 | 78.03 | 37.72 |
| SegNet [2] | ResNet34 | 37.24 | 54.27 | 62.79 | 47.78 |
| GCBNet [32] | ResNet34 | 60.44 | 72.70 | - | - |
| CoSwin Transformer [6] | ResNet34 | 61.28 | 75.99 | **79.75** | 72.57 |
| SWEG-Net [29] | ResNet34 | 60.67 | 74.25 | 75.69 | 72.86 |
| MSIF-Net [29] | ResNet34 | 59.31 | 72.73 | 74.71 | 70.85 |
| **Ours (D3FNet)** | ResNet34 | **63.16** | 75.75 | 76.67 | **77.99** |

To further investigate this issue, we conduct qualitative visual comparisons in the next subsection. Additionally, to evaluate the model's robustness and generalization, we also test D3FNet on the CHN6-CUG dataset. The CHN6-CUG dataset is also a widely used high-quality benchmark in the road extraction domain, known for its finer-grained annotations that effectively reduce the impact of missing labels on model evaluation.

Table 2 summarizes the performance of various models on this dataset. Specifically, the results for U-Net [28], U-Net++ [31], DeepLabV3+ [5], D-LinkNet34 [30], SWEG-Net [29], and MSIF-Net [29] are cited from Zhao et al. [29], while the performance of TransUNet [4], Swin Transformer [18], SegNet [2], GCBNet [32], and CoSwin Transformer [6] is referenced from Chen et al. [6] and Zhu et al. [32].

The results clearly demonstrate that D3FNet outperforms D-LinkNet34 across all evaluation metrics. Specifically, it achieves an IoU of 63.16%, representing an improvement of over 5% compared to D-LinkNet34. The F1-score reaches 75.75%, which is 6.49% higher than that of D-LinkNet. Precision increases to 76.67%, while Recall improves significantly to 77.99%, surpassing D-LinkNet's 66.21% by 11.78%. These results indicate a stronger ability to detect actual road regions and a notably lower miss rate.

Compared with more advanced models such as SWEG-Net and MSIF-Net, D3FNet also shows strong performance in Recall, achieving 77.99% compared to 72.86% for SWEG-Net and 70.85% for MSIF-Net. In terms of IoU, D3FNet reaches 63.16%, outperforming SWEG-Net at 60.67% and MSIF-Net at 59.31%. The F1-score of D3FNet reaches 75.75%, which is higher than SWEG-Net at 74.25% and MSIF-Net at 72.73%. CoSwin Transformer achieves the highest F1-score of 75.99%, slightly surpassing D3FNet. Although CoSwin Transformer also attains the highest Precision at 79.75%, D3FNet still performs better than SWEG-Net and MSIF-Net, whose Precision scores are 75.69% and 74.71%, respectively. In summary, D3FNet demonstrates excellent overall performance on the CHN6-CUG dataset. Its improvements in IoU and Recall validate its robustness and enhanced capability in extracting fine-grained road details under complex urban scenarios.

### 4.4.2. Qualitative Analysis

To further evaluate the effectiveness of our proposed model, we conducted a qualitative comparison with D-LinkNet34. Due to the unavailability of qualitative results or open-source implementations for some of the recent models such as SWEG-Net and MSIF-Net, we present the qualitative comparison between our proposed D3FNet and the well-established baseline D-LinkNet34, which shares a similar architecture and has demonstrated competitive performance in quantitative evaluation. Figure 5 and Figure 6 illustrate the visual comparisons on the DeepGlobe and CHN6-CUG datasets respectively. In the visualization, red dashed lines denote narrow road segments that are clearly visible in the satellite images but are omitted in the Ground Truth (GT), these areas are labeled as non-road in the GT. Blue dashed lines represent these "non-road" areas successfully identified by the models. The results demonstrate that D3FNet accurately detects many narrow roads omitted in the labels, while D-LinkNet performs poorly in these regions. This explains why D3FNet's Precision score is relatively low: many correctly identified roads are treated as false positives due to their absence in the GT, increasing the denominator of Precision.

In Figure 6, we analyze six representative images from the CHN6-CUG dataset, the left three images depict suburban satellite scenes and the right three show urban roads. D3FNet consistently delivers more stable and precise road extraction than D-LinkNet across both suburban and urban scenarios. Specifically, D3FNet successfully detects narrow roads present in the labels but missed by D-LinkNet, such as am100765_sat, am100797_sat, and hk100240_sat, and accurately extracts roads omitted in the annotations, such as am100863_sat and sh100748_sat. Moreover, in bj100521_sat, D3FNet correctly avoids classifying internal pathways within a campus as roads, as these paths are not truly drivable. This differs from other cases where D3FNet identifies narrow roads omitted in the GT; here the model's predictions align with the actual road network, demonstrating that it not only recovers missing labels but also avoids false positives in areas that are not real roads.

Overall, D3FNet exhibits strong capabilities in preserving road connectivity, recovering occluded or narrow road segments, and maintaining robustness in densely built urban scenes. The enhanced performance can be attributed to the differential attention mechanism, which effectively suppresses background noise and highlights salient road features, and the dual-stream decoder, which facilitates comprehensive feature integration and accurate reconstruction.

### 4.5. Ablation Study

We conducted ablation experiments on the CHN6-CUG dataset to evaluate the individual and combined contributions of the Differential Attention Dilation Extraction (DADE) module and the Dual-stream Decoding Fusion Mechanism (DDFM). The result of D-LinkNet is reported by Zhao et al. [29]. Compared quantitative results are shown in Table 3.

Incorporating the DADE module leads to significant improvements on the CHN6-CUG dataset, with IoU reaching 62.49%, F1-score increasing to 74.86%, precision achieving 79.88% and recall reaching 73.68%. When combined with DDFM, IoU further improves to 63.16%, F1-score to 75.75%, and recall rises to 77.99%, demonstrating that their integration significantly boosts the model's ability to detect narrow roads and overall performance.

Table 3. Ablation results of Differential Attention Dilation Extraction (DADE) module and Dual-stream Decoding Fusion Mechanism (DDFM) on CHN6-CUG dataset.

| Model Configuration | IoU (%) | F1-score (%) | Precision (%) | Recall (%) |
|---|---|---|---|---|
| D-LinkNet (Baseline) | 57.56 | 69.26 | 72.61 | 66.21 |
| D-LinkNet + DADE | 62.49 | 74.86 | **79.88** | 73.68 |
| D-LinkNet + DADE + DDFM (D3FNet) | **63.16** | **75.75** | 76.67 | **77.99** |

## 5. Conclusion

In this work, we proposed D3FNet, a Dilated Dual-Stream Differential Attention Fusion Network specifically designed for the fine-grained segmentation of narrow roads in high-resolution remote sensing imagery. The motivation stems from real-world challenges in cooperative autonomous driving and remote perception systems, where narrow, occluded, and low-contrast road segments are critical yet remain difficult to detect accurately. Through the integration of three core innovations—(1) the Differential Attention Dilation Extraction (DADE) module, (2) the Dual-stream Decoding Fusion Mechanism (DDFM), and (3) an optimized multi-scale dilation strategy—D3FNet effectively balances local structural precision with global semantic understanding. The DADE module enhances subtle road features while suppressing background noise, the DDFM alleviates gradient interference and facilitates fine-to-coarse information fusion, and the (1, 3, 5, 9) dilation configuration mitigates gridding artifacts to maintain road continuity.

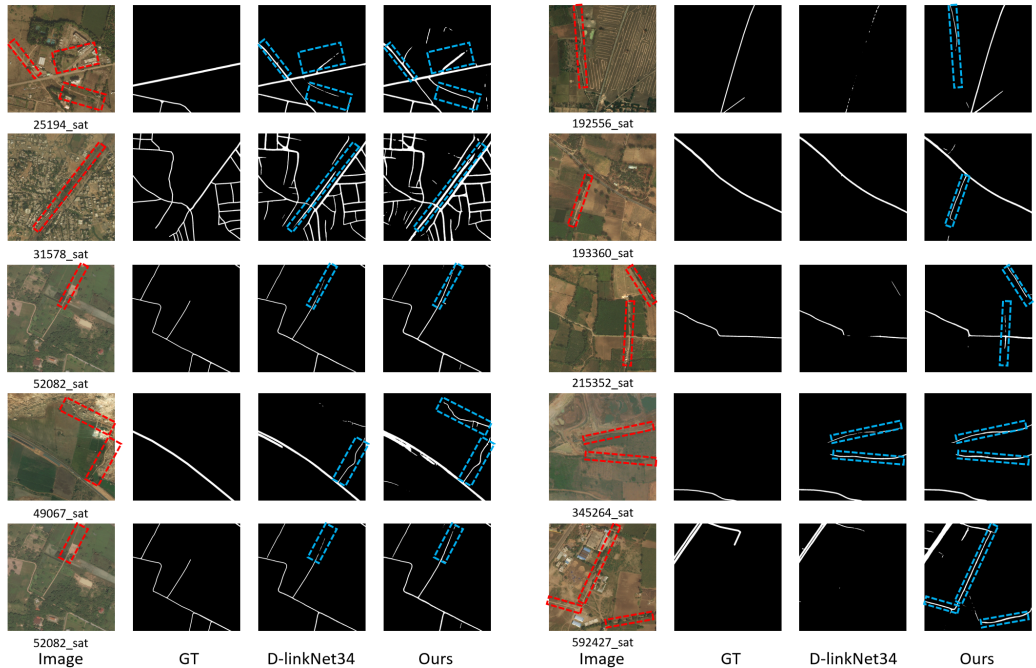

Figure 5. Narrow road extraction comparison on DeepGlobe dataset.

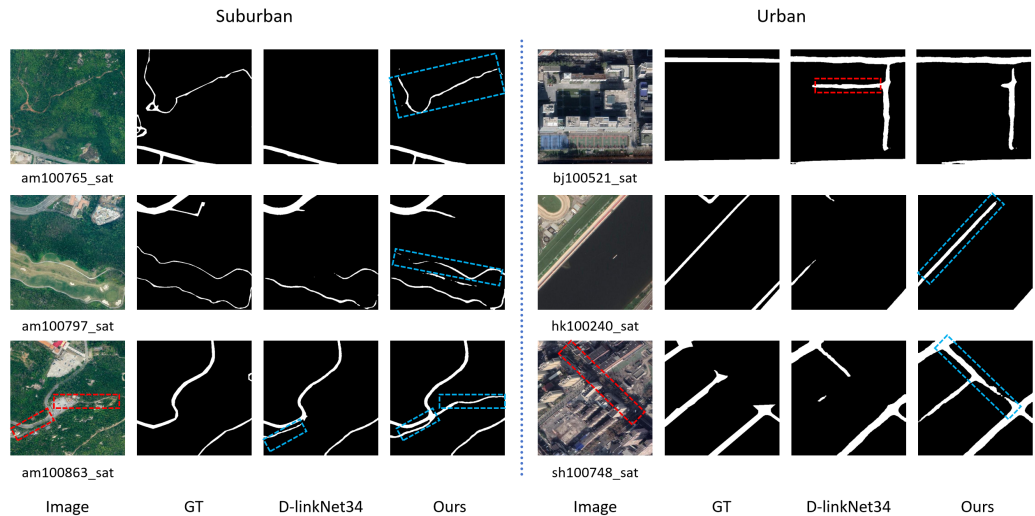

Figure 6. Narrow road extraction comparison on CHN6-CUG dataset.

Comprehensive experiments conducted on the DeepGlobe and CHN6-CUG datasets demonstrate that D3FNet outperforms state-of-the-art baselines in terms of IoU, Recall, and F1-score, particularly in cases involving fragmented and low-visibility road segments. The ablation studies further confirm the complementary benefits of combining differential attention with dual-stream decoding. These results suggest that D3FNet offers a solution for narrow road structure extraction in complex environments, providing valuable support for HD map generation, path planning, and autonomous navigation in cooperative perception systems. Future work will investigate D3FNet's computational cost, efficiency, and generalization to diverse environments to better assess practical applicability.

## Acknowledgements

This work was supported by the Kooperatív Technológiák Nemzeti Laboratórium (National Laboratory of Cooperative Technologies, KTNL) under project 2022-2.1.1-NL-2022-00012, funded by the Ministry of Culture and Innovation through the National Research, Development and Innovation Fund. Additional support was provided by the Hungarian National Research, Development and Innovation Office (NKFIH OTKA), Grant No. K139485. Project no. TKP2021-NVA-02 has been implemented with the support provided by the Ministry of Culture and Innovation of Hungary from the National Research, Development and Innovation Fund, financed under the TKP2021-NVA funding scheme.